%% file: main.tex
\documentclass[letterpaper, 10pt, conference]{ieeeconf}      

\IEEEoverridecommandlockouts 

\usepackage{times}
\usepackage{tabularx}
\usepackage{subcaption}
\usepackage{multirow,multicol, array}
\usepackage[font={small}]{caption}   
\usepackage{graphicx,dblfloatfix}
\usepackage{wrapfig}
\usepackage{siunitx}
\usepackage{soul}

\let\labelindent\relax
\usepackage{enumitem}

\usepackage{amsmath,amsthm,amssymb,amsfonts}
\usepackage[titlenumbered,ruled]{algorithm2e}
\usepackage{textcomp}
\usepackage{acronym}
\usepackage{balance}
\usepackage{mdwmath}
\usepackage{bm}
\usepackage{mdwtab}
\usepackage{array}
\usepackage{eqparbox}
\usepackage{cite}
\usepackage{verbatim}
\usepackage{amssymb}

\usepackage{color}
\usepackage{xcolor}
\usepackage{psfrag}
\usepackage{epsfig}
\usepackage{url}

\usepackage{epstopdf}
\usepackage{booktabs}
\usepackage{blindtext}

\usepackage{physics}

\usepackage[colorinlistoftodos,prependcaption,textsize=tiny]{todonotes}

\renewcommand{\emph}{\textit}

\newtheorem*{lemma*}{Lemma}

\newtheorem*{problem*}{Problem}

\makeatletter
\newcommand\fs@spaceruled{\def\@fs@cfont{\bfseries}\let\@fs@capt\floatc@ruled
    \def\@fs@pre{\vspace{5\baselineskip}\hrule height.8pt depth0pt \kern2pt}%
    \def\@fs@post{\kern2pt\hrule\relax}%
    \def\@fs@mid{\kern2pt\hrule\kern2pt}%
    \let\@fs@iftopcapt\iftrue}
\def\maketag@@@#1{\hbox{\m@th\normalfont\normalsize#1}}
\makeatother

\IEEEoverridecommandlockouts
\graphicspath{{images/}}

\newcommand{\mycomment}[1]{}

\title{A Novel Lockable Spring-loaded Prismatic Spine \\to Support Agile Quadrupedal Locomotion 
}
	
\author{Keran Ye, Kenneth Chung and Konstantinos Karydis
\thanks{The authors are with the Dept. of Electrical and Computer Engineering, University of California, Riverside. 
Email: {\{kye007, kchun048, karydis\}@ucr.edu}.
}
\thanks{
We gratefully acknowledge the support of NSF \# CMMI-2046270, ARL \# W911NF-18-1-0266, and a UC MRPI Award. Any opinions, findings, and conclusions or recommendations expressed in this material are those of the authors and do not necessarily reflect the views of the funding agencies.}
}
	
\begin{document}

\maketitle
\thispagestyle{empty}
\pagestyle{empty}


\begin{abstract}
This paper introduces a way to systematically investigate the effect of compliant prismatic spines in quadrupedal robot locomotion. We develop a novel spring-loaded lockable spine module, together with a new Spinal Compliance-Integrated Quadruped (SCIQ) platform for both empirical and numerical research. Individual spine tests reveal beneficial spinal characteristics like a degressive spring, and validate the efficacy of a proposed compact locking/unlocking mechanism for the spine. Benchmark vertical jumping and landing tests with our robot show comparable jumping performance between the rigid and compliant spines. An observed advantage of the compliant spine module is that it can alleviate more challenging landing conditions by absorbing impact energy and dissipating the remainder via feet slipping through much in cat-like stretching fashion.
\end{abstract}

\section{Introduction}
Owing to developments in actuators, embedded single-board computers and perception units, quadrupedal robots have become more versatile in various \emph{agile} locomotion tasks~\cite{ijspeert2014biorobotics,hwangbo2019learning,pandala2022robust} including rapid running~\cite{park2017high}, aggressive jumping~\cite{rudin2021cat,nguyen2019optimized}, fast stepping~\cite{bledt2020extracting}, and other acrobatic maneuvers~\cite{katz2019mini,chignoli2021humanoid,bledt2020extracting,lee2019robust}. 
A range of existing quadrupedal platforms~\cite{guizzo2019leaps,bouman2020autonomous,bellegarda2022robust,smith2022legged,katz2019mini,bledt2018cheetah,hutter2017anymal,lee2020learning,ma2020bipedal,blackman2016gait,semini2015design,ijspeert2014biorobotics,gan2022energy,zucker2011optimization} adopt a single rigid body (SRB) design and extend the overall morphological degree-of-freedoms (DoFs) by employing 3-DoF legs. 
This approach philosophy can result in modeling representations that can be directly embedded into robot controllers and estimators. 

Further extending morphological DoFs may be the key to achieve more agile locomotion~\cite{koutsoukis2016passive,li2020trotting,fisher2017effect,zhang2016effects,liu2022systematic,ye2023evaluation,casarez2018steering}. However, adding more actuated joints within the leg assemblies could be rather expensive~\cite{caporaletwisting}. 
On the other hand, biological inspiration from mammals~\cite{zhang2014effect,schilling2010function,alexander1985elastic,schilling2011evolution} suggests the potential of applying more spinal DoF(s), in contrast to a rigid body and in addition to the current approaches regarding leg DoFs, to enable novel control strategies for more agile maneuverability. 

Existing biological evidence has demonstrated the varied spinal functions of bending and twisting in vertebrate studies~\cite{bennett2001twisting}. In turn, such studies have motivated several bio-mimetic multiple-DoF spine models and prototype designs that have been found capable of stabilizing gaits~\cite{zhao2012embodiment,hustig2016morphological,takuma2010facilitating,eckert2015comparing,kani2011effect,eckert2018towards,khoramshahi2013piecewise,li2020trotting}, enhancing speed~\cite{folkertsma2012parallel,kani2011effect,eckert2018towards,pusey2013free,khoramshahi2013piecewise,bhattacharya2019learning}, and promoting efficiency~\cite{eckert2018towards,haomachai2021lateral,kani2011effect,khoramshahi2013piecewise,bhattacharya2019learning}. 
Further abstractions of the spinal morphology to 1-DoF spines~\cite{khoramshahi2013benefits,chong2021coordination,chen2017effect,duperret2017empirical,koutsoukis2016passive,fisher2021optimal,culha2011quadrupedal,pouya2017spinal,yesilevskiy2018spine,phan2020study,fisher2017effect,caporaletwisting,ye2021modeling} have also gained attention owing to their simpler mathematical representation to investigate more essential functionality. 
Spines with a revolute joint in either sagittal, frontal, or transverse planes have been widely studied numerically and empirically for their contribution to motion regularity~\cite{chong2021coordination,culha2011quadrupedal,duperret2017empirical,fisher2017effect,khoramshahi2013benefits,koutsoukis2016passive,pouya2017spinal,caporaletwisting,ye2021modeling} and transport efficiency~\cite{chen2017effect,chong2021coordination,kani2013comparing,phan2020study,pouya2017spinal,yesilevskiy2018spine}. 
Although spine bending also implies the linear change in vertebrae morphology, spines embodied with a prismatic joint have been explored only scarcely, with most of the work~\cite{ye2021modeling,koutsoukis2016passive,fisher2017effect,fisher2021optimal} conducted in numerical settings.

\begin{figure}[!t]
	\vspace{4pt}
	\centering
\includegraphics[trim={0cm 0cm 0cm 0cm},clip,width=0.985\linewidth]{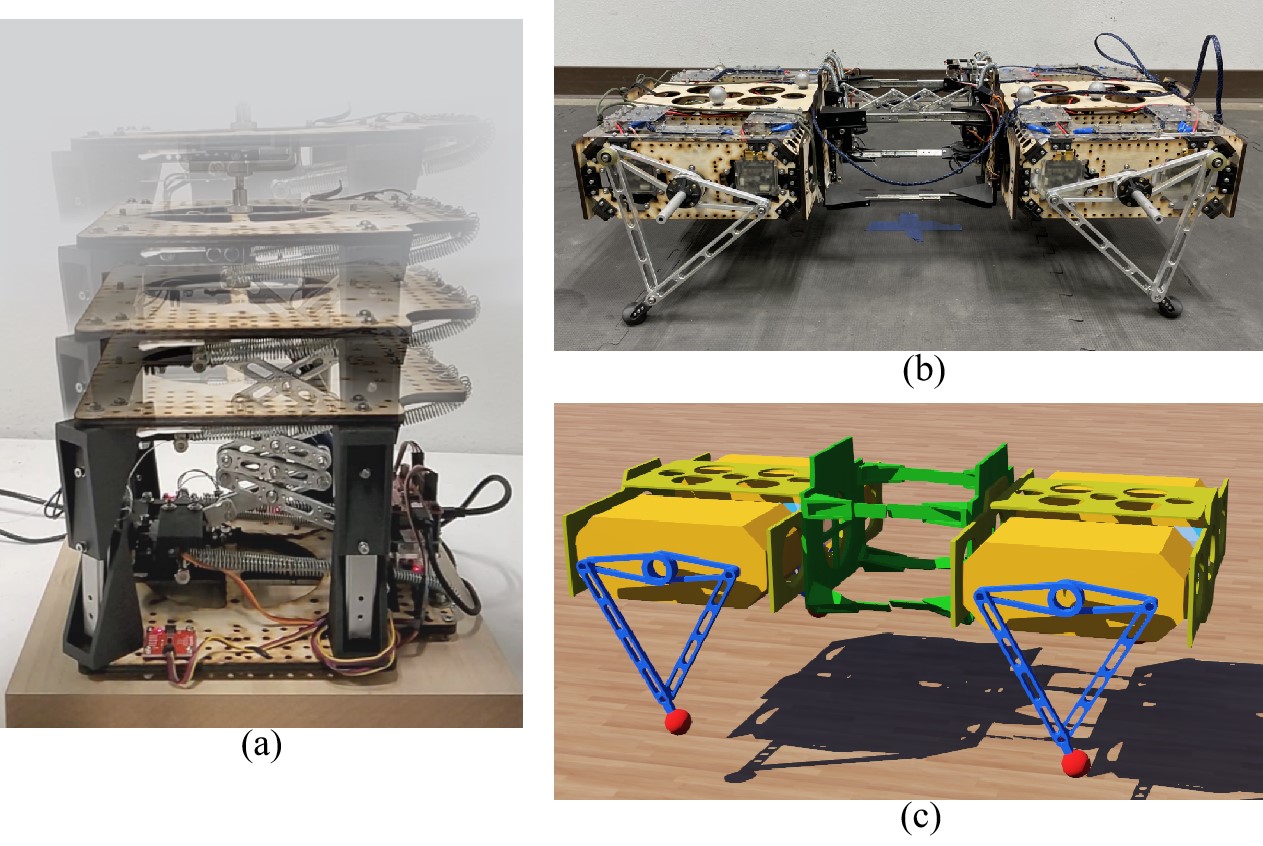}
	\vspace{-4pt}
	\caption{\textbf{The SCIQ Robot.} (a) The novel lockable spring-loaded prismatic spine module proposed in this work. (b) The SCIQ robot prototype with the lockable compliant prismatic spine in place. (c) A Webots simulator model we created for our robot with identical properties as in the physical robot yet simplified visual rendering. }	\label{fig:SCIQ_robot}
	\vspace{-18pt}
\end{figure}

Vertebrate studies also suggest that the spine actuation source is compounded with actuators (from muscles) and compliance (from ligaments and tendons)~\cite{schilling2010function,alexander1985elastic,schilling2011evolution}. 
Revolute joints can be readily deployed into actuators and compliant components compactly, and can allow significant morphological change between stretching and contracting a robotic spine. 
Thus, several efforts have considered revolute joints for spinal DoFs, shown to lead to overall efficiency improvement~\cite{chen2017effect,fisher2017effect,kani2013comparing,phan2020study,pouya2017spinal}, high-speed stability~\cite{culha2011quadrupedal,duperret2017empirical,khoramshahi2013benefits,pouya2017spinal} during agile motion, and acrobatic maneuvers~\cite{caporaletwisting}. 
On the other hand, prismatic joints' engineering design is less straightforward under the same space requirements for compactness. 
Direct deployment of a prismatic spine framework (like linear rails) in the sagittal axis is risky as other parts of the robot may clash with extruding rail parts. Actuation of any compliance integration in this axis is also challenging without compromising the large shape adjustment (we offer more details in Section~\ref{sec:Spine}).


Besides the technological challenges and opportunities around (compliant) spine design, modeling and control, it is also important to develop quadrupedal platforms that can afford integration of different spine prototypes interchangeably. Doing so can help quantify each design's effectiveness, strengths, and limitations. One platform that has this feature is the small-scale robot Lynx~\cite{eckert2015comparing}. Nevertheless, spine exchangeability remains an open area for the quasi-direct-drive (QDD) family of quadrupedal robots~\cite{kenneally2016design,wensing2017proprioceptive} on a comparatively larger scale.

The work underlying this paper aims to provide a means to systematically study the effect of compliant prismatic spines in quadrupedal robot locomotion. 
%
We seek to offer design insights for one such spine prototype and a new QDD legged platform that can integrate different spine modules interchangeably and with minimum engineering effort (Figure~\ref{fig:SCIQ_robot}). 
\emph{The major technical result of this paper concerns a novel spring-loaded lockable spine module, together with a new Spinal Compliance-Integrated Quadruped (SCIQ) platform}. We focus on the design, analysis and experimental testing of the spine on its own and while embedded into our new quadrupedal robot. In detail, our work contributes: 
1) a novel compact and light-weight prismatic spine module with independent controller and sensors;
2) an automatic spine locking/unlocking mechanism with tiny servos triggering for large spinal force; 
3) a low-cost quadrupedal robot with convenient hardware interfaces for spine modules;  
and 
4) a controller framework with distributed coordination with spine modules and a synchronized interface to the simulator.

\section{Quadrupedal Platform}
We begin by discussing our new robot and its various hardware and software traits first. Then, the description and analysis for the novel spine are provided in the next section.

\subsection{Overview}
\begin{figure}[!t]
	\vspace{0pt}
\includegraphics[trim={0cm 0cm 0.5cm 0cm},clip,width=\linewidth]{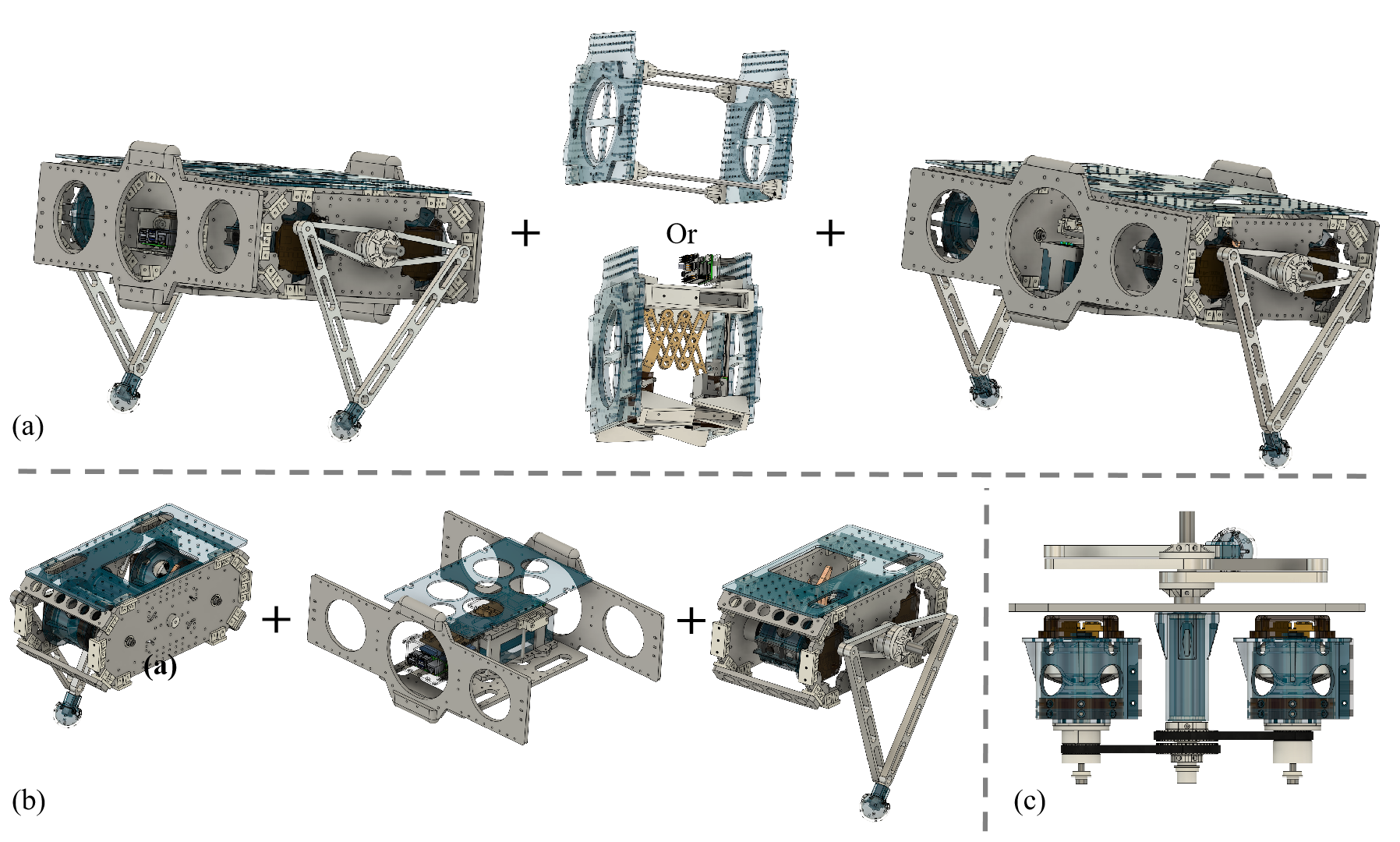}
	\vspace{-15pt}
	\caption{\textbf{The SCIQ Robot Design.} (a) The robot assembly comprises one half-body, a spine module, and the other half-body. (b) Each half-body consists of two 2-DOF leg modules and a half-body trunk. (c) Each leg module adopts a flat-symmetric servo layout and links to leg limbs via timing belts and co-axial shafts. }
	\label{fig:Robot_Design}
	\vspace{-6pt}
\end{figure}

The SCIQ robot is a member of the QDD legged robot family where existing quadrupedal platforms~\cite{kenneally2016design,kau2019stanford,seok2012actuator} have offered several design principles and recommendations we follow in this work (see Section~\ref{sec:Robot_Hardware}). 
In our previous work~\cite{ye2021modeling} we introduced a quadrupedal model to investigate agile maneuvers with a lock-enabled compliant prismatic spine. In this work, we develop physical prototypes of both spine and quadruped and study spinal dynamics embodied in legged robots for agile locomotion. SCIQ is designed with an eye to facilitating integration of different spines interchangeably as well as being (cost-)efficient in terms of fabrication, maintenance, and upgrades.  
%
As such, the SCIQ robot is composed of one exchangeable spine module and two power-independent half-bodies, with each half-body including two leg modules, as shown in Figure~\ref{fig:Robot_Design}a and~\ref{fig:Robot_Design}b.

\input{Tables/property_summary}

\subsection{Hardware} \label{sec:Robot_Hardware}
\textbf{Actuation:} 
There exist several successfully-deployed actuators such as direct-drive electric motors~\cite{kenneally2016design,kau2019stanford,seok2012actuator}, series elastic actuators (SEA)~\cite{hutter2017anymal}, and hydraulic ones~\cite{semini2015design}. Here we choose a brushless DC (BLDC) motor (T-Motor MN5212 kv340) with an off-the-shelf planetary gear set (0.5 module, 6:1 reduction) and an open-source BLDC motor control board (MJBOTS Moteus r4.5). Our custom servo module (Figure~\ref{fig:Robot_Design}c) is fitted into a 3D-printed (Markforged Mark 2 and Onyx material) plastic enclosure. The actuator can achieve position-velocity-torque control with high torque density (Table~\ref{table:property_summary}). 

\textbf{Leg and body:}
Each leg module accommodates two servos and co-axial shafts (that link to leg limbs) in a hip assembly in a flat-symmetric manner to balance the mass about the hip center (Figure~\ref{fig:Robot_Design}c). 
The pre-tensioned timing belts not only transmit motion between the leg limbs and servos but also serve as a failure buffer to protect servos from extreme impact upon legs during agile maneuvers. 
The body frame contains the controller pack and the battery pack with direct access to electric connection and disconnection. It is bridged to the leg modules through simple bolting between its cross frames and the leg's hip assembly, in an effort to facilitate integration of additional servos for ab/ad joints in future work. 
All frame plates are made of birch wood and linked with 3D-printed plastic connectors to offer a good tradeoff between weight, rigidity, and budget. 

\begin{figure}[!t]
	\vspace{5pt}
\includegraphics[trim={0cm 0cm 0cm 0cm},clip,width=\linewidth]{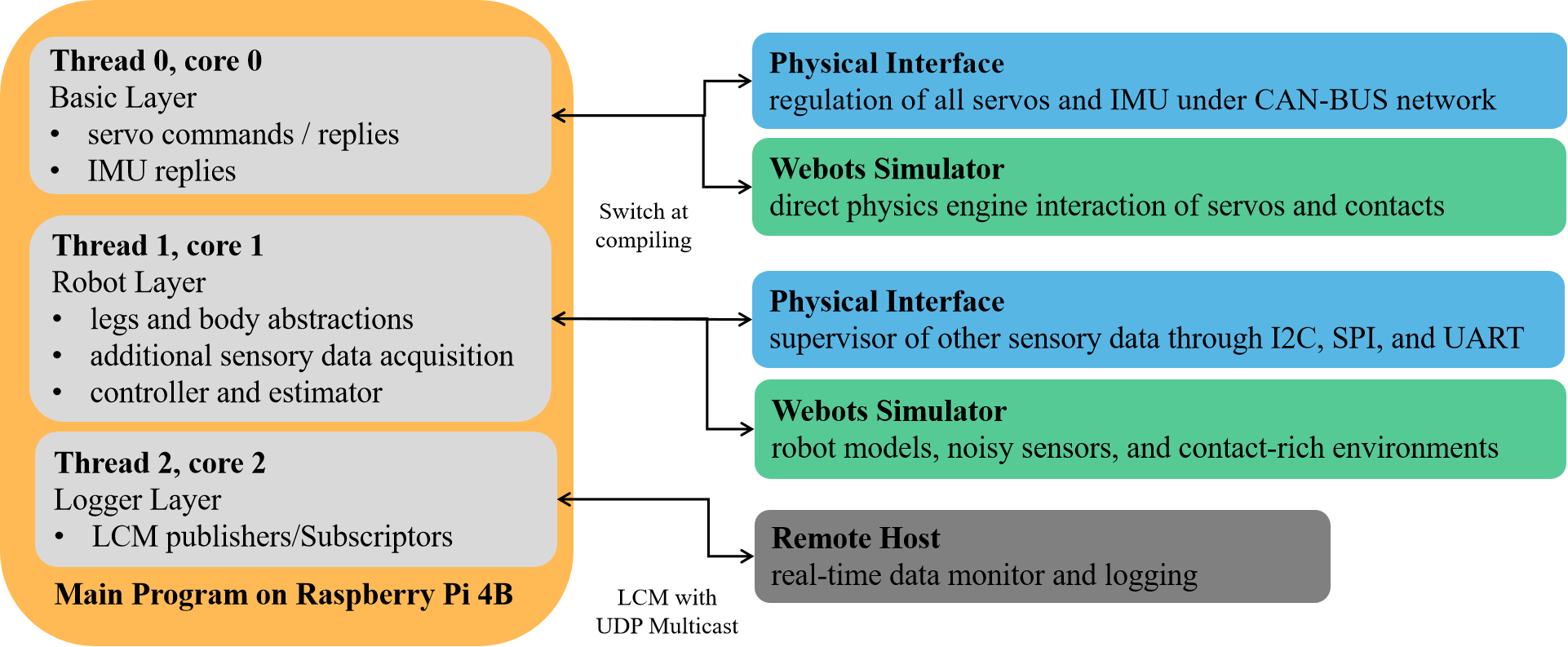}
	\vspace{-12pt}
	\caption{\textbf{Software Architecture.} The main program (yellow) translates low-level commands for servos, computes estimation and control input for the robot, and handles data loggers in multiple threads (light grey). The physical interface (blue) encodes all servo commands and replies through CAN-BUS. The simulation interface functions similarly to the physical interface except in a Webots simulator. The remote host (dark grey) serves to monitor the robot behavior and log necessary data. }\label{fig:Software_Framework}
	\vspace{-18pt}
\end{figure}

\subsection{Software}
The controller pack consists of a main computer (Raspberry Pi 4B+), a servo communication board (MJBOTS Pi3hat r4.4), and a LED cooling fan. 
The main computer runs a Preempt-RT patched Linux kernel to enable a soft real-time execution of the whole controller program. It works with the servo communication board to regulate all servos through four CAN-FD BUS lines, with each line responsible for one leg module.
A minimal finite state machine (FSM) is used to allow for online scheduling of different locomotion tasks.

Figure~\ref{fig:Software_Framework} demonstrates the architecture of the software framework.
The multithreaded main program executes on the main computer for all necessary calculations and interacts with the physical interface for all information exchange between the servos (through the servo communication board) and other units (like the controller of the compliant spine module).
Simulation is synchronized with the main program in parallel and is carried out in Webots with the direct use of the provided physical engine for the sake of more realistic legged behaviors in contact-rich contexts.
It is worth noting that two different CPU architectures are used for Webots simulation (on a laptop with X86-64 CPUs) and real robot (Raspberry Pi running on ARM CPUs). To alleviate the coding adaptation workload,  
we carefully developed the simulation interface as analogous to the physical interface as possible, and the switching between these two interfaces can be done at compiling time with a single flag set.

The communication between the robot control pack and the spine module is distributed as two custom nodes on the Lightweight Communications and Marshalling (LCM)~\cite{huang2010lcm} network, in which way any other spine module can be effortlessly linked to the main controller node with its own LCM node following certain message principles.


\section{Spine Module} \label{sec:Spine}

\subsection{Overview} \label{sec:Spine_Overview}
Many existing works abstract the biological evidence of the spine bending~\cite{alexander1985elastic,bennett2001twisting} during locomotion with one or more revolute joints~\cite{khoramshahi2013benefits,chen2017effect,duperret2017empirical,koutsoukis2016passive,fisher2021optimal,culha2011quadrupedal,pouya2017spinal,yesilevskiy2018spine,phan2020study,fisher2017effect,caporaletwisting,chong2021coordination,ye2021modeling}. 
Yet, the resulting considerable change in body length along the sagittal axis also implies the possibility of a prismatic joint interpretation, which has been verified as being effective for facilitating some quadrupedal dynamical gaits~\cite{fisher2021optimal,fisher2017effect}. 
Notwithstanding the direct deployment of any linear actuator as the actuated prismatic spine, \emph{passively compliant spine actuation} remains an attractive alternative because the compliance may introduce high-density energy storage in lightweight components and high-frequency response through mechanical feedback~\cite{yesilevskiy2018spine,koutsoukis2016passive}.
The locking/unlocking functionality for a spine can enable different locomotion modes in various situations~\cite{eckert2015comparing}. The spine may be locked to act like a single rigid body when a robot requires accurate motion with low speed, and unlocked to be passively compliant for less precise but more agile locomotion.

The aforesaid evidence signifies the importance of developing such a compliant prismatic spine with lockable features. Thus, we propose our compact spine design with detailed descriptions of its mathematical model (Section~\ref{sec:Spine_Model}), the locking/unlocking strategy (Section~\ref{sec:Spine_Lock}), and its physical realization (Section~\ref{sec:Spine_Prototype}).

\begin{figure}[!h]
	\vspace{-6pt}
\includegraphics[trim={0cm 0cm 0cm 0cm},clip,width=\linewidth]{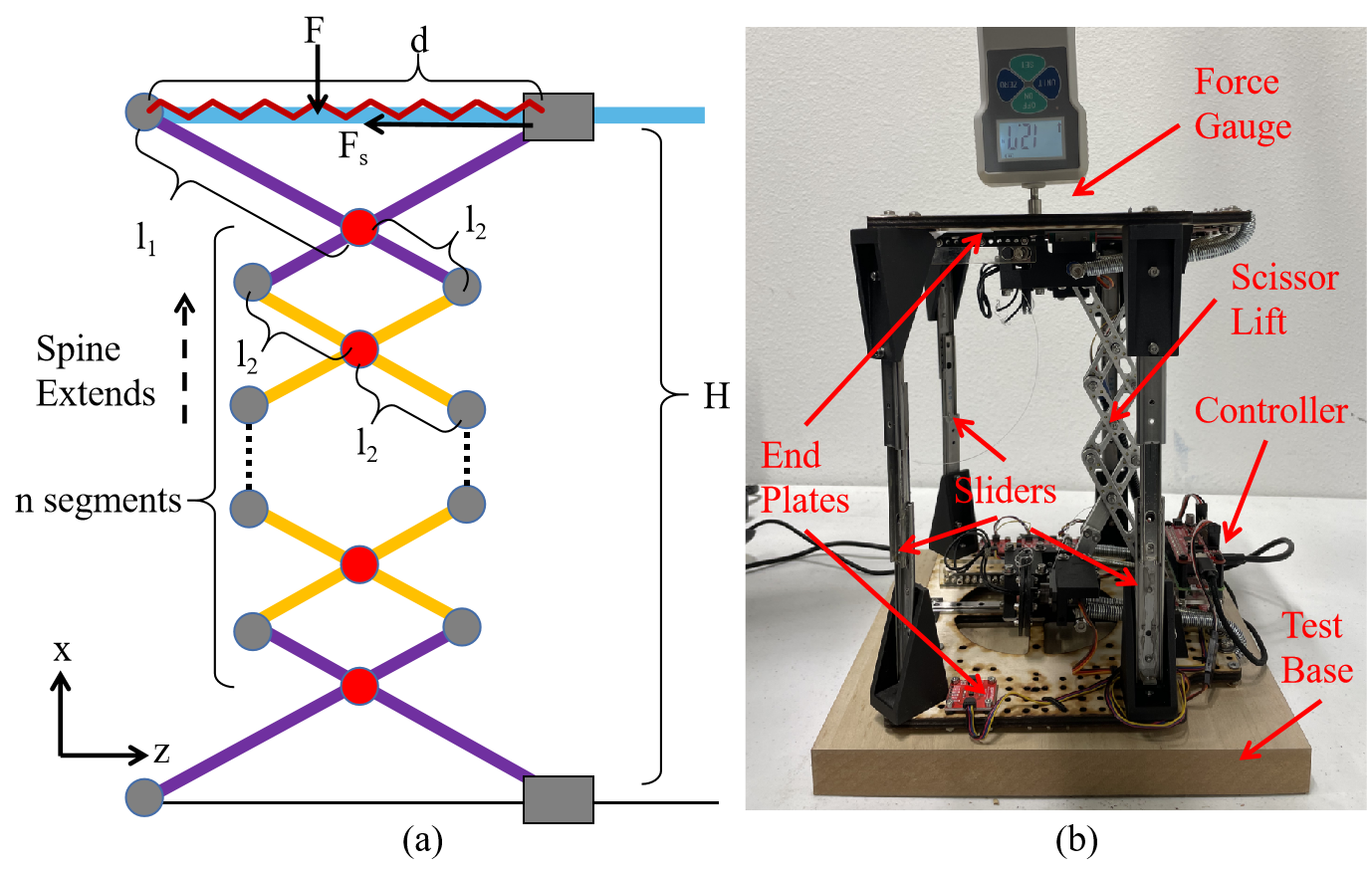}
	\vspace{-15pt}
	\caption{\textbf{Spine Model and Prototype.} (a) the spine model with $n$ scissor segments: scissor lift links are marked in yellow and purple, end plates with linear rail/carriage group in blue and black, revolute joint as grey and red dots, and prismatic joints as grey blocks. (b) the spine prototype and the corresponding experimental setup.}\label{fig:spring_model_vs_prototype}
	\vspace{-12pt}
\end{figure}

\subsection{Model Analysis} \label{sec:Spine_Model}
Figure~\ref{fig:spring_model_vs_prototype}a shows the underlying model for the proposed spine inspired by the scissor lift that is widely used for different loading applications~\cite{takesue2016scissor} while keeping itself as compact as possible with multiple scissor segments. 
By transforming the motion in the $x$ axis (spine extending direction) to the $z$ axis (actuation direction), a larger spine length change can be achieved with a shorter actuation distance through $n$ scissor segments. 
A non-even pattern is applied for the spinal scissor lift structure to give more freedom to the geometric adjustment over the spine extension and the actuation span.

\begin{figure*}[!t]
	\vspace{6pt}
\includegraphics[trim={0cm 0cm 0cm 0cm},clip,width=\linewidth]{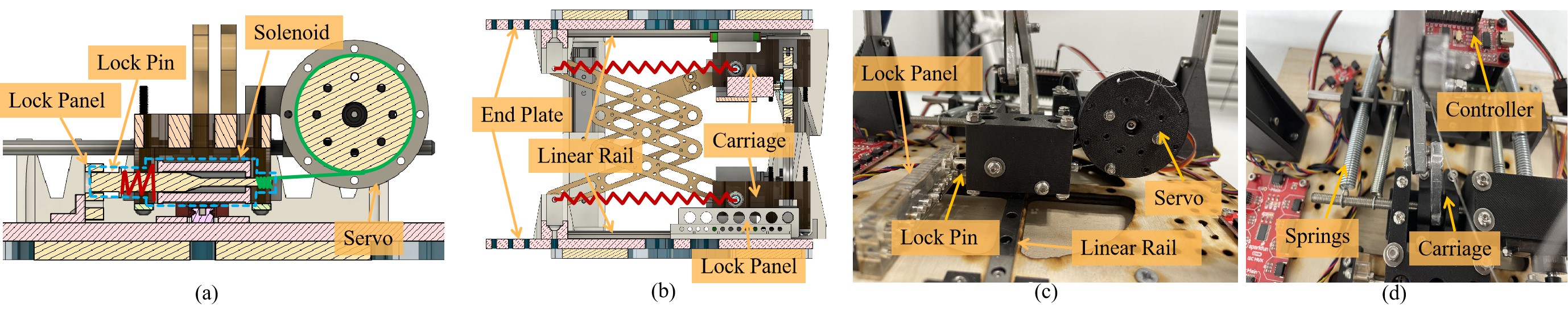}
	\vspace{-18pt}
	\caption{\textbf{Spine End Plate and Locking/Unlocking Mechanism.} 
 (a) The locking/unlocking mechanism design: a solenoid (at zero volts, circled in dashed blue lines) is pulled by a tiny servo through a string (green) and rebounds with a spring (red).
 (b) The end plate configuration: a linear rail sits along a lock panel and accommodates a carriage unit and springs (red) on each end plate.
 (c) The locking/unlocking prototype.
 (b) The end plate prototype with a controller pack.
 }\label{fig:spine_end_plate_lock}
	\vspace{-18pt}
\end{figure*}

We assume that all scissor segments and end plates are massless, and all joints are frictionless. The scissor segments attached to the end plates consist of links with length $(l_1+l_2)$ and the rest scissor segments in the middle have link length  $(2 l_2)$. This leads to the simplified geometric-based model 
\begin{align}
    H = n' \sqrt{4 l_{1}^{2} - d^{2}}, \:
    n' = 1 + ( n - 1 ) \frac{l_2}{l_1}\enspace,
    \label{equ:spine_geo}
\end{align}
where $H$ is the spine extension length and $d$ is the actuation span length. $n'$ is a transformed scissor segment count that will also be used in force analysis. Due to the massless assumption above, we can ignore the inertia, thus yielding 
\begin{align}
    F = \frac{2}{n'} (H - H_0 \sqrt{ \frac{4 l_1^{2} n'^{2} / H_0^{2} - 1}{4 l_1^{2} n'^{2} / H^{2} - 1} }  ) K_s, \: H \le H_0\enspace,
    \label{equ:spine_force}
\end{align}
where $K_s$ is the expansion spring constant and $H_0$ is the spine extension length when the compression spring is at rest length. Note that $H_0$ may be beyond the reach of $H$ if the spring is pre-tensioned at the shortest actuation span. We separate $F$ into linear, $F_{lin}$, and nonlinear, $F_{nonlin}$, terms as    
\begin{align}
    F &= F_{lin} + F_{nonlin} \\ \nonumber
    F_{lin} &= \frac{2}{n'} H  K_s, \,
    F_{nonlin} = - \frac{2}{n'} H_0 \sqrt{ \frac{4 l_1^{2} n'^{2} / H_0^{2} - 1}{4 l_1^{2} n'^{2} / H^{2} - 1} } K_s
    \label{equ:spine_force_lin_nonlin}
\end{align}
where $F_{lin}$ is linear in regard to spine length $H$. $F_{nonlin}$ diminishes with $H$ near 0 and reaches $-\frac{2}{n'} H_0 K_s$ as $H \rightarrow H_0$.

Interpretation of~\eqref{equ:spine_geo} and~\eqref{equ:spine_force} suggests nonlinear geometric and force relationships, while the force curve can be approximated linearly when the spine length is short (i.e. spine in compressed status). 
On the other hand, the force curve has one peak before $H$ reaches $H_0$ and its value ($H_{peak}$) can be obtained by numerically solving a 6th-degree polynomial equation.
Thus, the spine maximum length $H_{max}$ is recommended to be designed near $H_{peak}$ because $F$ will decrease rapidly toward 0 after the spine extends over ${H_{peak}}$.

Another inspiring observation is that \emph{the proposed spine model will function similarly to a degressive spring} when ${H < H_{peak}}$, that is, becomes easier to compress as it is further compressed. Such a tendency makes it easier to restore elastic energy when the spine is forced back to its shortest length by any external force. When it comes to utilizing elastic energy by extending the spine length, this feature helps exert more spinal force to propel the robot's body instead of the force degression characterized by a normal compression spring model.

\mycomment{
\begin{figure}[!t]
	\vspace{6pt}
\includegraphics[trim={0cm 0cm 0cm 0cm},clip,width=\linewidth]{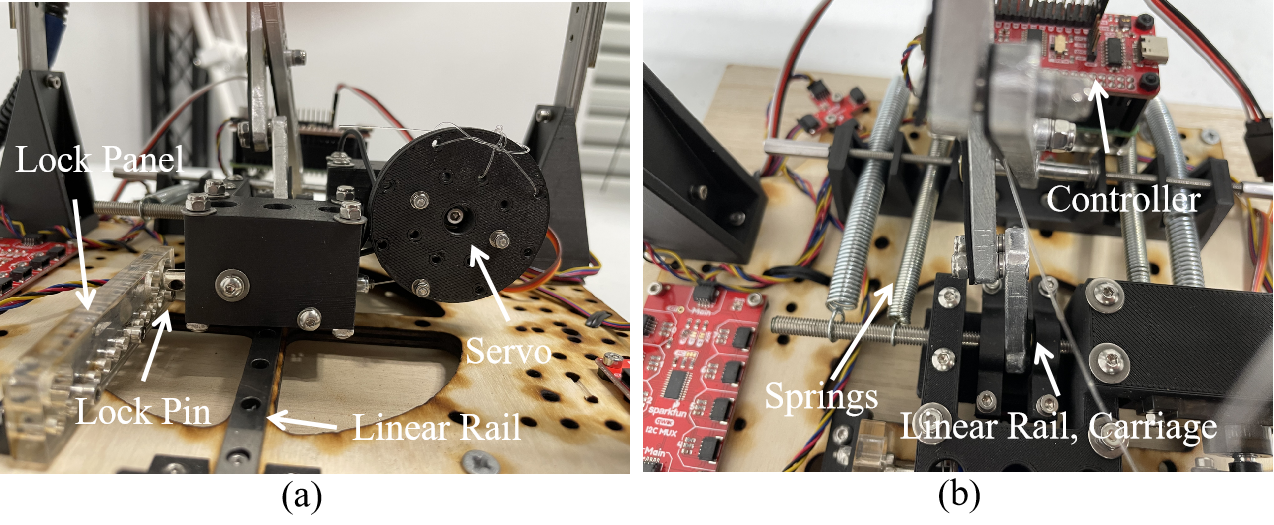}
	\vspace{-18pt}
	\caption{\textbf{Spine End Plate and Locking/Unlocking Mechanism.} (a) the locking/unlocking mechanism is accomplished by a solenoid (at zero volts) and tiny servo.  (b) the end plate accommodates controller pack and linear rail/carriage group to tension springs.}\label{fig:spine_end_plate_lock}
	\vspace{-18pt}
\end{figure}
}

\subsection{MOSFET-Like Locking/Unlocking Mechanism} \label{sec:Spine_Lock}
To lock the spine while it is moving, one can turn to the friction-based method (like a bike brake) or conflict-based method (like a door lock). The friction-based method requires more power for generating fiction to counteract the body motion and is not suitable for compact deployment. The conflict-based method relies on the material stiffness and toughness to handle great impact on the conflict and is not limited any less by the the compact space requirements. 

\input{Tables/spine_parameters}

To this end, we choose the conflict-based method and propose a locking/unlocking mechanism based on a compact solenoid-servo system (Figure~\ref{fig:spine_end_plate_lock}a and~\ref{fig:spine_end_plate_lock}c).
This system is mounted on a carriage, together forming a prismatic joint sliding along the linear rail and springs (Figures~\ref{fig:spine_end_plate_lock}b and~\ref{fig:spine_end_plate_lock}d). 
The solenoid provides a metal pin that is pushed forward by the attached spring and inserts a portion into a locking hole on the lock panel. This way, the prismatic joint is locked on the linear rail and subsequently stops the spine motion through the scissor lift transform. When unlocking is needed, the tiny servo rotates its string wheel and pulls out the pins.
However, considerable spine elastic force is necessary to act along with leg actuation and thrust the whole robot forward during agile maneuvers.
Such a condition will expose the locking pin to a great amount of friction when it is pulled by the servo and it introduces risk to compromise the hardware. 

\begin{figure*}[!t]
\vspace{6pt}
\centering
\includegraphics[trim={2.5cm 1.2cm 2.9cm 0cm},clip,width=\linewidth]{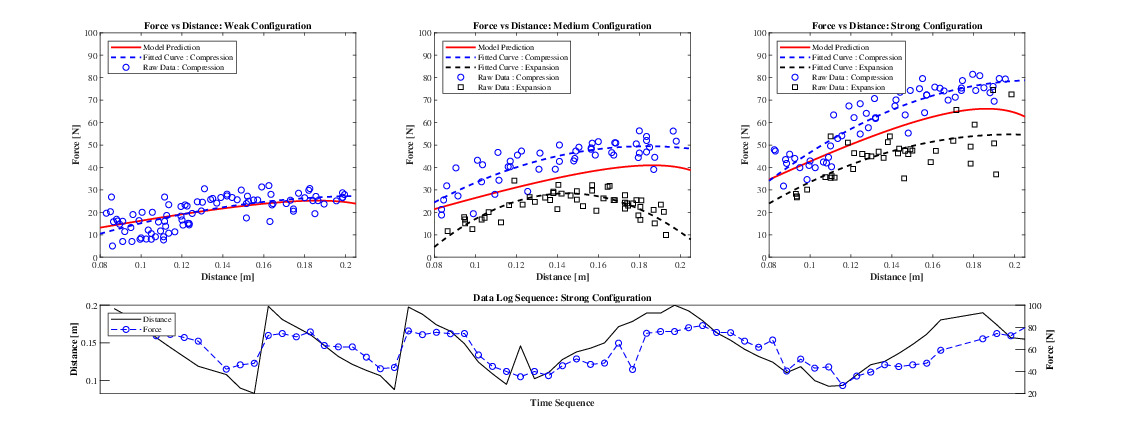}
\vspace{-15pt}
\caption{\textbf{Spine Characteristics Test.} The upper three graphs show results for weak, medium, and strong spring configurations: the model prediction curve (red), data from experiments (blue circle for spine compression, black square for spine extension), and fitted curves (blue dashed line for spine compression, black dashed line for spine extension). Before visualization, the raw data were downsampled and filtered to remove outliers. The raw data are fitted with order-2 polynomial curves. 
The lower graph shows results from four consecutive spine compression-extension trials with distance data (black line) and force data (circle-dashed blue line). 
An external force is applied by hand to regulate the spine motion varying between trails such that it moves slower in the last two trails. 
(Figure best viewed in color.)}
\label{fig:Spine_characteristics_test}
\vspace{-6pt}
\end{figure*}

\begin{figure*}[!t]
\vspace{0pt}
\centering
\includegraphics[trim={0cm 0cm 0cm 0cm},clip,width=\linewidth]{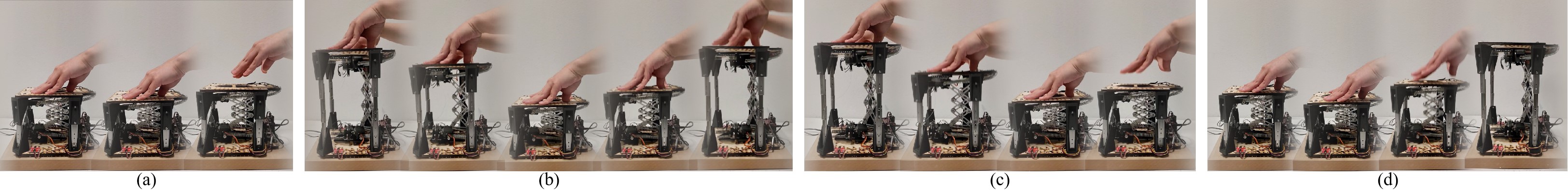}
\vspace{-15pt}
\caption{\textbf{Spine Locking/Unlocking Tests.} (a) The spine is pre-locked and commanded to stay locked. (b) The spine is pre-unlocked and commanded to stay unlocked. (c) The spine is pre-unlocked and commanded to be locked at the shortest length. (d) The spine is pre-locked and commanded to be unlocked at the shortest length when the press is detected. The sequences were sampled at 5Hz. The supplemental video file demonstrates these tests more clearly. }
\label{fig:Spine_lock_test}
\vspace{-15pt}
\end{figure*}

To address this challenge, we propose a perception-based locking/unlocking algorithm based on the following logic. 
\begin{itemize}
    \item When the pin is locked at a certain spine length $H_{lock}$, it can be unlocked only when unlocking is commanded \emph{and} an external press happens to release the pin from close contact with the lock hole and then allow the servo to pull out the pin.
    \item The external press is captured by a modified cumulative sum control chart (CUSUM)~\cite{page1954continuous} edge detection algorithm at the spine locking length $H_{lock}$. 
    \item When the pin is unlocked, any lock command will allow the servo to release the spring-loaded pin into the nearest lock hole to freeze the spine.
\end{itemize} 
As such, the locking/unlocking mechanism can act like a MOSFET transistor (which can control a much larger current with minimum input current) to release and retain significant elastic energy with small locking/unlocking actuation power.


\subsection{Prototype} \label{sec:Spine_Prototype}

Figure~\ref{fig:spring_model_vs_prototype}b shows the lockable compliant spine prototype. 
The whole spine module weighs 1.2 kg, which is heavier than the 0.6 kg rigid spine (Table~\ref{table:property_summary}), yet it can exert up to 80 N spine force from the springs. 
The main structure comprises two end panels linked by four miniature sliders that confine the spine length range. 
Each end panel is laser cut (VLS 3.60) from birch wood board and contains a miniature linear rail with a locking/unlocking system powered by a tiny SG90S servo. 
The scissor lift structure is spring-loaded and bolted along the linear rails on the two end panels.

The controller pack includes a single-board computer (Raspberry Pi Zero 2 Wireless) and a servo control hat (SparkFun Servo pHAT) that connects to a pair of distance sensors (SparkFun Distance Sensor VL53L4CD) for edge detection mentioned in Section~\ref{sec:Spine_Lock}. The spine length is estimated by the average of the two distance sensor measurements and falls back to one measurement if the other fails which has been occasionally observed during experiments.

Table~\ref{table:spine_paramsters} presents the scissor lift structure design parameters with specific values under the careful cross-study of the slider limitation, available spring geometry, and $H_{peak}$ calculation (Section~\ref{sec:Spine_Model}). 
It is evident that our spine module can expand up to twice its shortest length while providing increasing thrust. The expanding scale is similar to the leg limbs (Table~\ref{table:property_summary}) and thus could be of considerable benefit to enlarge the stride length and expedite gait speed~\cite{yesilevskiy2018spine,koutsoukis2016passive}. 

\section{Results and Discussion} 



\subsection{Spine Characteristics Test} \label{sec:Spine_Test_Characteristics}

Figure~\ref{fig:spring_model_vs_prototype}b depicts the testbed used to perform spine characterization experiments.
The spine was standing upright with the bottom end plate fixed to a flat wooden table base. A force gauge was used to apply force at the center of the top end plate while recording force measurements.

\begin{figure*}[!t]
	\vspace{-6pt}
\includegraphics[trim={0cm 0cm 0cm 0cm},clip,width=\linewidth]{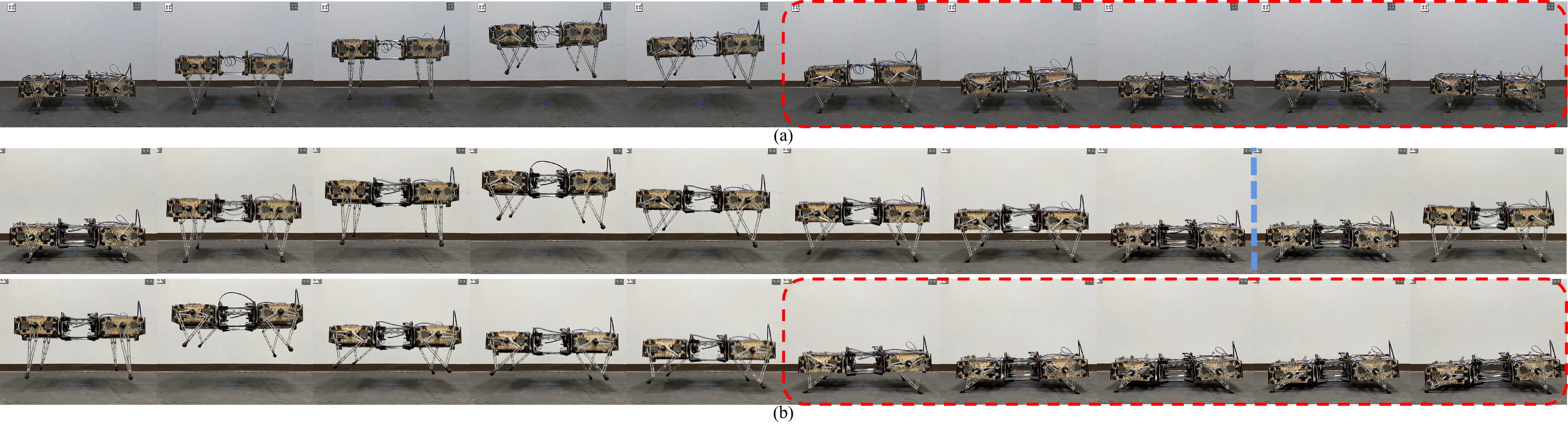}
	\vspace{-15pt}
	\caption{\textbf{Jumping Test Sequence.} (a) One jumping and landing trial with the rigid spine. (b) Two consecutive jumping and landing trials with the spring-loaded spine. These are extending over the middle and bottom row panels. A blue vertical dashed line denotes end of the first trial and initiation of the second. The first jump results in similarly mild landing conditions like in (a) yet without any body bouncing. The second jump exposes more challenging landing situations which can be overcome owing to the spring-loaded spine. Critical landing sections are marked in red dashed boxes to facilitate a side-by-side comparison of the robot behavior with the rigid (top) and with the compliant (bottom) spine equipped. The robot always faced leftwards. The supplemental video demonstrates these tests clearly.}
	\label{fig:jump_test_sequence}
	\vspace{-15pt}
\end{figure*}

We performed tests with two types of linear extension springs. Both have the same rest length $d_0=0.096$\;m but different spring constants (soft: $K_{s,soft} = 224$\;N/m; stiff: $K_{s,stiff} = 364$\;N/m).
Combinations of these springs result in three distinctive compliant configurations, namely weak (four soft springs), medium (four stiff springs), and strong configuration (four soft springs and four stiff springs). 
For each configuration, we performed 20 trials with spine compression-extension routines and collected data through the spine's onboard computer and the force gauge. 

Figure~\ref{fig:Spine_characteristics_test} (upper panels) show the results following testing of the above three compliance configurations. Results suggest that our proposed simplified model offers a satisfactory prediction for the force-length relation.   
The force-length curve during spine compression (or extension) is with an observable positive (or negative) offset from the predicted curve, mainly due to the sliders' friction acting along (or against) the spinal force. 
Note that the spine extension routine for the soft configuration was not able to be properly recorded due to the serious effect of sliders' friction and clogging which, however, can be overcome by the medium and strong configurations.
It is also evident that the strong configuration outperforms the rest with the spinal force peak close to the longest length.
The bottom panel of Figure~\ref{fig:Spine_characteristics_test} shows the force-length curves during consecutive trials along the timeline and visualizes the spine module's degressive-spring-like feature that the spinal force and length are positively correlated.



\subsection{Spine Locking/Unlocking Verification} \label{sec:Spine_Test_Lock}
The verification experiment of the lock/unlocking mechanism was conducted on the same setup as the characteristic test (Section~\ref{sec:Spine_Test_Characteristics}).
Figure~\ref{fig:Spine_lock_test} demonstrates four different situations the locking/unlocking mechanism may encounter during locomotion described below. 
\begin{itemize}
    \item When the spine is currently locked and the robot commands the spine to stay locked (Figure~\ref{fig:Spine_lock_test}a).
    \item When the spine is currently unlocked and the robot commands the spine to stay unlocked so that the former is free to move (Figure~\ref{fig:Spine_lock_test}b).
    \item When the spine is currently unlocked and the robot commands the spine to lock when it is pressed to its shortest length (Figure~\ref{fig:Spine_lock_test}c).
    \item When the spine is currently locked and the robot commands the spine to unlock when it is pressed at its shortest length (Figure~\ref{fig:Spine_lock_test}d).
\end{itemize}

Results show that the proposed locking/unlocking mechanism produced the desired actions for each situation. In addition, the proposed strategy can respond to fast spine motion and large spinal force in a real-time manner. 
On the other hand, a small bounce-back was empirically observed when the spine was locked (Figure~\ref{fig:Spine_lock_test}a and~\ref{fig:Spine_lock_test}c), mainly because of small wobbling of the solenoid pin across its motion axis under large spinal force. 
We expect to resolve this issue in future work by fabricating a custom pin and prismatic joint with linear bearing to minimize wobbling motion and enhance the overall unit stiffness.



\subsection{Robot Vertical Jumping and Landing Test} \label{sec:Spine_Test_Jump}
To evaluate the efficacy of the passive prismatic spine into high-impact aggressive quadruped locomotion, we focused in this paper on vertical jumping and landing. 
Based on our findings about spine properties (Section~\ref{sec:Spine_Test_Characteristics}), we chose the strong compliance configuration for our compliant spine module. Comparisons are made against a rigid spine module; both can be equipped in an interchangeable manner to our SCIQ robot.  

We performed 20 trials of vertical jumping and landing for each spine module with a similar battery level. All servos were limited to 70\% of their peak torque. The SCIQ robot was connected to a local WIFI network shared with a remote host which was used only for collection of proprioceptive data from servos and onboard IMU through LCM. 
The remote host was also saving the robot's posture information provided by a VICON motion capture system in the lab. At the same time, we recorded all motions of the robot with a high-speed action camera for post-analysis.

We investigated whether or not spinal compliance can affect the landing phase, particularly when the robot is not landing properly with the legs because of body tilting and varied contact times among feet upon impact. 
Figure~\ref{fig:jump_test_sequence}a shows one of the successful jumping tests for the SCIQ robot with the rigid spine.
Despite repeatable jumping performance overall, it was occasionally observed that when the rigid spine was used, the timing belt had some slippage in the case that the landing posture was not ideal and the legs did not compensate properly for the large ground reaction forces that could cause the robot's body bounce as shown in the red-marked region in Figure~\ref{fig:jump_test_sequence}a.  
On the other hand, no such observation was made with the compliant spine, which was compressed to some extent to mitigate the impact from the legs when an unfavored landing posture happened. Figure~\ref{fig:jump_test_sequence}b captures this effect in the red-circled segment, and shows that our spring-loaded spine can absorb the impact energy which is 1) partially stored as the spine's elastic energy and 2) dissipated for the rest through the front feet slipping like a cat's stretching under the spine's degressive spring property (Section~\ref{sec:Spine_Model}).

\begin{figure}[!t]
	\vspace{3pt}
\includegraphics[trim={0cm 0cm 0cm 0cm},clip,width=\linewidth]{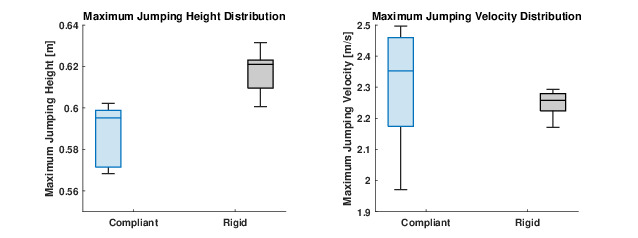}
	\vspace{-15pt}
	\caption{\textbf{Jumping Characteristics.} The left and right box charts show the distribution of the maximum jumping height and velocity respectively. Compliant spine's results are in blue and rigid in grey. }
	\label{fig:jump_test_characteristics}
	\vspace{-20pt}
\end{figure}

Figure~\ref{fig:jump_test_characteristics} exhibits side-by-side the results of both spine modules in terms of vertical jumping height and maximum velocities. The graphs suggest that the two spine modules share similar jumping performances, on average. 
The robot with the compliant spine experienced a little smaller jumping height due to the extra weight of the spine compared to the rigid one, yet it was capable of producing larger peak velocities.
Notwithstanding the fact that the rigid spine delivers more consistent jumping behaviors, the compliant spine enables more aggressive motion in exchange of precision, a feature which can serve well in some agile locomotion tasks.

\section{Conclusions}
The paper contributes a novel spring-loaded lockable spine module together with a new quadrupedal 
platform called SCIQ, in an effort to promote spinal compliance research for agile legged locomotion both empirically and numerically. 
Obtained results reveal the advantageous spinal properties similar to a degressive spring, and justify the effectiveness of the proposed compact locking/unlocking mechanism for the spine. 
Physical experiments with the robot suggest similar jumping performance between the rigid and compliant spine modules, yet giving the compliant spine enjoys better performance in more challenging landing situations. 
Future directions of research include 1) the improvement of the locking/unlocking unit for better stiffness, 2) the robot's self-manipulation with the spine lockable feature, and 3) a more comprehensive effectiveness study of various spine designs for different locomotion gaits.





\end{document}

%% file: Tables/property_summary.tex
\begin{table}[!th]
\vspace{0pt}
    \caption{Summary of Key SCIQ Robot Properties} 
    \vspace{-10pt}
    \label{table:property_summary}
    \begin{center}
    \renewcommand{\arraystretch}{1.5}
    \resizebox{0.99\columnwidth}{!}{
    \begin{tabular}{p{3.35cm}>{\centering}p{2.075cm}>{\centering}p{2.025cm}}
        \toprule
        \textbf{Properties} & Symbol & Value  \\ 
        \midrule
        \midrule
        Half-body, Battery Mass
        & $m_{half}$, $m_{batt}$
        & 4.6 kg, 0.7 kg \\ 
        Upper, Lower Limb Mass
        & $m_{ulimb}$, $m_{llimb}$ 
        & 45 g, 55 g \\ 
        Upper, Lower Limb Length
        & $l_{ulimb}$, $l_{llimb}$ 
        & 0.1 m, 0.2 m \\ 
        Rigid Spine Mass
        & $m_{rspine}$
        & 0.6 kg \\ 
        Rigid Spine Length
        & $l_{rspine}$ 
        & 0.23 m \\ 
        Compliant Spine Mass
        & $m_{cspine}$ 
        & 1.2 kg 
        \\ 
        Compliant Spine Length
        & $l_{cspine}$ 
        & 0.12-0.24 m 
        \\ 
        \midrule
        Motor, Servo Mass
        & $m_{motor}$,  $m_{servo}$
        & 0.21 kg, 0.44 kg \\ 
        Motor, Shaft Peak Torque
        & $\tau_{motor}$,  $\tau_{shaft}$
        & 0.92 Nm, 9.2 Nm \\ 
        Overall, Gear, Belt Reduction
        & $r$, $r_{gear}$, $r_{belt}$
        & 10:1, 6:1, 1.67:1 \\ 
        
        \bottomrule
        
    \end{tabular}
    }
    \end{center}
    \footnotesize{$\:$ The shaft is one of the co-axial shafts and $l_{cspine}=H + \delta H$ (Table~\ref{table:spine_paramsters}).}
    \vspace{-25pt}
\end{table}


%% file: Tables/spine_parameters.tex
\begin{table}[!t]
\vspace{6pt}
    \caption{Compliant Spine Design Parameters} 
    \vspace{-9pt}
    \label{table:spine_paramsters}
    \begin{center}
    \renewcommand{\arraystretch}{1.5}
    \begin{tabular}{p{3.2cm}>{\centering}p{2.25cm}>{\centering}p{1.9cm}}
        \toprule
        \textbf{Parameters} & Symbol * 
        & Values  \\ 
        \midrule
        \midrule
        Number of scissor segments
        & $n$,
        & 3 \\ 
        Scissor lift limb length
        & $l_{1}$, $l_{2}$ 
        & 0.03 m, 0.06 m \\ 
        Spine Extending Length
        & $H$ 
        & 0.08-0.20 m \\ 
        Spring Extending Length
        & $d$ 
        & 0.06-0.10 m\\ 
        
        \bottomrule
        
    \end{tabular}
    \end{center}
    \footnotesize{* Symbols in this table are defined in Figure~\ref{fig:spring_model_vs_prototype}a. The actual spine length $l_{cspine}$ (Table~\ref{table:property_summary}) is $H$ plus the installment offset $\delta H = 0.04 m$.}
    \vspace{-18pt}
\end{table}
